\documentclass[sigconf]{acmart-me}

\usepackage{booktabs} 
\usepackage{url}
\usepackage{color}
\usepackage{enumitem}
\setcopyright{rightsretained}

\acmDOI{}

\acmISBN{}

\acmConference[MediaEval'20]{Multimedia Evaluation Workshop}{December 14-15 2020}{Online} 
\acmYear{2020}
\copyrightyear{}

\acmPrice{}

\begin{document}
\title{Multi-modal Ensemble Models for Predicting Video Memorability}

\author{Tony Zhao\textsuperscript{1}, Irving Fang\textsuperscript{1}, Jeffrey Kim\textsuperscript{1}, Gerald Friedland\textsuperscript{1}}
\affiliation{\textsuperscript{1}University of California, Berkeley\\}
\email{tonyzhao@berkeley.edu, irvingf7@berkeley.edu, jeffjohn3@berkeley.edu, fractor@eecs.berkeley.edu}

\renewcommand{\shortauthors}{Zhao et al.}
\renewcommand{\shorttitle}{Predicting Media Memorability}

\begin{abstract}
Modeling media memorability has been a consistent challenge in the field of machine learning. The Predicting Media Memorability task in MediaEval2020 is the latest benchmark among similar challenges addressing this topic. Building upon techniques developed in previous iterations of the challenge, we developed ensemble methods with the use of extracted video, image, text, and audio features. Critically, in this work we introduce and demonstrate the efficacy and high generalizability of extracted audio embeddings as a feature for the task of predicting media memorability.
\end{abstract}

\maketitle

\section{Introduction and Related Work}
\label{sec:intro}
The MediaEval2020 Predicting Media Memorability Task aims to predict how memorable videos are. The dataset consists of videos with short-term and long-term memorability scores, pre-extracted images and video features, and other information described in detail in \cite{overviewPaper}. Participants predict the probability that each video will be remembered in both the short (after a few minutes) and long (after 24-72 hours) term. 

Image-based features extracted from pre-trained convolutional neural networks (CNNs) have been demonstrated as useful in predicting the memorability of videos in the dataset. These features in combination with semantic-embedding and image captioning models have also proved effective for predicting memorability scores \cite{cohendet}. Video-based features, such as C3D and I3D, have also recently been considered in the study of video memorability \cite{gibis}. Finally, the best performing models of the 2019 iteration of the Predicting Media Memorability challenge utilized ensemble models with the above-mentioned features \cite{azcona2019predicting}.
\begin{table*}[ht]
  \caption{Spearman's rank correlation coefficient (SRCC) of features for short-term (ST) and long-term (LT) memorability over 5 runs trained and evaluated on training set (80-20 train-validation split on 590 videos)}
  \label{tab:features_st}
  \begin{tabular}{cccllll}
    \toprule
    Modality & Feature & Model  & Mean (ST) & Variance (ST) & Mean(LT) & Variance (LT) \\
    \midrule
    video & \textbf{C3D} & SVR & 0.152 & 0.081 & 0.076 & 0.059 \\
    image & \textbf{ResNet152} & SVR & 0.233 & 0.096 & 0.133 & 0.066\\
    image & VGG & SVR & 0.167 & 0.058 & \textbf{0.144} & 0.092\\
    image & LBP & SVR & 0.139 & 0.072 & 0.024 & 0.067\\
    audio & \textbf{VGGish} & Bayesian Ridge & \textbf{0.246} & \textbf{0.017} & 0.059 & \textbf{0.026}\\
    text & \textbf{GloVe} & GRU & 0.200 & 0.029 & 0.104 & 0.091\\
  \bottomrule
\end{tabular}
\end{table*}
\section{Approach}
\label{sec:approach}
The training set comprises of 590 short videos (1-8 seconds long) with 2-5 human-annotated captions each. Another development set of 410 additional videos was provided later, but was not used as we discovered the quality of annotated memorability scores to be worse than that of the original set. Each video has corresponding short-term and long-term memorability scores, which were adjusted using methods outlined in previous work \cite{khosla} \cite{cohendet}, namely the following update functions to calculate decay rate $\alpha$ and memorability $m_T$ at duration $T$
$$\alpha \leftarrow 
\frac{\sum^N_{i=1}\frac{1}{n^{(i)}} \sum^{n^{(i)}}_{j=1} \log(\frac{t^{(i)}_j}{T})[x^{(i)}_j - m^{(i)}_T] }
{\sum^N_{i=1}\frac{1}{n^{(i)}} \sum^{n^{(i)}}_{j=1}[ \log(\frac{t^{(i)}_j}{T})]^2}
$$
$$
m_T^{(i)} \leftarrow
\frac{1}{n^{(i)}} \sum^{n^{(i)}}_{j=1}[x^{(i)}_j - \alpha \log(\frac{t_j^{(i)}}{T})]
$$
where we have $n^{(i)}$ observations for video $i$ given by $x^{(i)} \in \{0,1\}$ and $t^{(i)}_j$ where $x_j=1$  implies that the repeated video was correctly detected when shown after time $t_j$. The average $t$ was 74.96, so we adjusted the memorability score for each video to be $m_{75}$ after 10 iterations with $\alpha$ converging to -0.0264.

Models were then trained on a 80-20 training-validation split on video id. Since concatenating multiple multi-modal features resulted in extremely high-dimensional feature vectors which were difficult to train with, we trained various models (Support Vector Regressor, Bayesian Ridge Regressor, Linear Models) on each individual feature independently.

We also explored several prediction aggregation functions for when a video has multiple embeddings of a specific feature, such as several different model predictions due to multiple captions for a given video. While previous work generally defaulted to a simple average, we discovered that taking the median value performed consistently better than either mean, max, or min. In the opposite scenario, where a video has no feature such as a soundless video for audio-based models, we defaulted to using the average of all predicted memorability scores.

We noticed high variance in the performance of the models, which we attributed to the relatively small dataset. Therefore we ran the feature models over 5 random seeds and took the best performing features of each modality (VGGish for audio, ResNet152 for image, C3D for video, GloVe for text). The predictions of these models were then used to perform grid-search over permutations of buckets of 5\% to calculate a weighted average as our final ensemble model.

\subsection{Audio Features}
Using VGGish \cite{vggish}, a pre-trained CNN model (trained on AudioSet), we extracted 128-dimensional embeddings for each second of video audio. Each video had on average had 5.6 embeddings extracted, with one soundless video having none. Bayesian Ridge Regressor provided the best performance on these features.

\subsection{Image/Video Features}
Local Binary Patterns (LBP), VGG \cite{vgg}, and Convolutional 3D (C3D) proved to have notable performance among the provided features. We discovered that Support Vector Regressor (SVR) \cite{SVR} models produced the best results with these features. 

We also extracted 8 equally spaced frames from each video, which were used for our own image-based feature extraction. The penultimate layer of a pre-trained ResNet-152 \cite{resnet152} (trained on ImageNet) was used to extract a 2048-dimensional feature vector. Similarly, we found that the best performance came from SVR models trained on the extracted features.

Following the methods outlined in \cite{azcona2019predicting}, pre-trained facial-emotion models were used to extract emotion-based features on the video frames. Ultimately, these features were not used in any of the final models, as only approximately a third of the videos had detectable faces as well as overall poor performance from all emotion-based models.

\subsection{Text Features}
Provided with several human-annotated captions for each video, we explored several different methods to extract semantic features. Past work suggested that simpler methods like bag-of-words could outperform more sophisticated methods in terms of both effectiveness and efficiency \cite{EURECOM+6062}. We vectorized the captions using bag-of-words before training with ordinary least squares, ridge, and lasso regression models. Bag-of-words vectorization and linear models performed the worst among our text-based approaches and were not used in the final model.

We tokenized the captions before extracting 300-dimensional GloVe \cite{pennington2014glove} word embeddings (trained on Wikipedia 2014 and Gigaword 5) to vectorize the caption tokens. These vectors were used as input to train a recurrent neural network with gated recurrent units (GRU) \cite{cho-etal-2014-learning}. Our final text-based model had 64 units in the initial GRU layer with a dropout of 0.8, followed by 4 dense layers with a dropout of 0.25 and ReLU activation, trained for 150 epochs with early stopping, a learning rate of 0.001, batch size of 64, and an Adam optimizer.

We also explored machine-generated captions \cite{autoCaption} to augment our text-based approaches. After generating captions for each video based on the first, middle, and last frames and mixing the generated captions with the human-annotated captions, we discovered that the performance improvement was insignificant. These automatic captions were not included in our final models.

\begin{table}[t]
  \caption{Short-term memorability ensemble models, Validation SRCC on 20\% of training set, Test SRCC on testing set}
  \label{tab:ensemble_st}
  \begin{tabular}{ccccccc}
    \toprule
    Model & C3D & ResNet152 & VGGish & GloVe & Valid & Test \\
    \midrule
    1 & 0.20 & 0.35 & 0.45 & 0.00 & 0.343 & \textbf{0.136} \\
    2 & 0.00 & 0.20 & 0.35 & 0.45 & 0.345 & 0.116 \\
    3 & 0.05 & 0.00 & 0.50 & 0.45 & \textbf{0.370} & 0.085 \\
    4 & 0.00 & 0.50 & 0.15 & 0.35 & 0.357 & 0.091 \\
    5 & 0.35 & 0.00 & 0.30 & 0.35 & 0.317 & 0.102 \\
  \bottomrule
\end{tabular}
\end{table}

\begin{table}[t]
  \caption{Long-term memorability ensemble models, Validation SRCC on 20\% of training set, Test SRCC on testing set}
  \label{tab:ensemble_lt}
  \begin{tabular}{ccccccc}
    \toprule
    Model & C3D & ResNet152 & VGGish & GloVe & Valid & Test \\
    \midrule
    1 & 0.00 & 0.40 & 0.15 & 0.45 & \textbf{0.289} & 0.012 \\
    2 & 0.55 & 0.10 & 0.00 & 0.35 & 0.192 & 0.076 \\
    3 & 0.00 & 0.55 & 0.45 & 0.00 & 0.118 & 0.044 \\
    4 & 0.25 & 0.35 & 0.20 & 0.20 & 0.168 & \textbf{0.077} \\
    5 & 0.30 & 0.05 & 0.00 & 0.65 & 0.201 & 0.056 \\
  \bottomrule
\end{tabular}
\end{table}

\section{Results and Analysis}
The notable features are seen in Table \ref{tab:features_st}, with the bolded features being selected for our final ensemble models, seen in Table \ref{tab:ensemble_st} and Table \ref{tab:ensemble_lt}.
Given the small dataset, we observe relatively high variance in performance. VGGish had the best performance, smallest variance, and thus presumably the best generalization for short-term memorability. However, VGGish features tended to perform poorly for predicting long-term memorability despite maintaining low variance.

The difference in validation and test Spearman's rank correlation in our final ensemble models are likely due in part to overfitting after performing grid-search over a relatively small dataset. In addition, we noticed that the quality of annotations between datasets were different, which may cause distribution differences in memorability scores and consequently poorer performance at test time.

\section{Discussion and Outlook}
Our main contribution is the demonstration that audio-based models perform well for predicting short-term memorability and generalize much more readily than other methods with the dataset. This may be in part due to the low dimensionality of the extracted audio embeddings. We reiterate that ensembling models of different modalities achieve the best performance as each model represents a different high-level abstraction of the data.

The size of the dataset was notably smaller than similar datasets for the task of predicting media memorability. Future work would ideally iterate on our findings for much larger datasets.

\newpage
\bibliographystyle{ACM-Reference-Format}
\def\bibfont{\small} 
\bibliography{reference.bib} 

\end{document}